\documentclass[11pt,twocolumn,journal]{IEEEtran}

\usepackage{amsmath,amssymb,bm,calc,epsfig,multicol,color,graphicx,psfrag}
\graphicspath{{fig/}}

\usepackage[nospace]{cite}
\usepackage{color}
\usepackage{subfig,float,multirow}
\usepackage{mdwtab}

\newcommand{\bigO}[0]{\mathcal{O}}

\newcommand{\x}{\mathbf{x}} 
\newcommand{\X}{\mathbf{X}} 

\newcommand{\w}{\mathbf{w}} 
\newcommand{\kk}{\mathbf{k}}  
\newcommand{\muw}{\mathbf{\mu_\w}} 
\newcommand{\noiseOut}{{\nu}} 

\newcommand{\indexData}{i}
\newcommand{\id}{\matr{I}}       

\def\K{{\mathbf K}} 
\def\kernel{{k}}
\def\fv{f} 
\def\fp{\vect{f}} 
\def\y{{y}} 
\def\yv{\vect{\y}} 
\def\tset{\mathcal{D}} 

\newcommand{\vect}[1]{\mathbf{#1}}     
\newcommand{\matr}[1]{\mathbf{#1}}     

\newcommand{\LABEQ}[1]{\label{eq:#1}}
\newcommand{\LABFIG}[1]{\label{fig:#1}}
\newcommand{\EQ}[1]{\eqref{eq:#1}}
\newcommand{\FIG}[1]{Fig.~\ref{fig:#1}} 

\renewcommand{\vect}[1]{{\boldsymbol{\mathbf{#1}}}} 
\newcommand{\mat}[1]{{\boldsymbol{\mathbf{#1}}}} 

\newcommand{\Kb}{\mat{K}}

\newcommand{\h}{\vect{h}}

\newcommand{\pmean}{\vect{\mu}}
\newcommand{\pcov}{\mat{\Sigma}}
\newcommand{\vt}[2]{\begin{bmatrix} #1 \\ #2  \end{bmatrix}}
\newcommand{\mtt}[4]{\begin{bmatrix} #1 & #2 \\#3 & #4  \end{bmatrix}}

\usepackage{url}
\usepackage{hyperref}
\hypersetup{
    pdftitle={Gaussian Processes for Nonlinear Signal Processing},
    pdfauthor={Fernando Perez-Cruz, Steven Van Vaerenbergh, Juan Jose Murillo-Fuentes, Miguel Lazaro-Gredilla and Ignacio Santamaria},
    breaklinks=true,
}

\begin{document}

\title{Gaussian Processes for Nonlinear Signal Processing}
\author{Fernando P\'erez-Cruz\thanks{Fernando P\'erez-Cruz and Miguel L\'azaro-Gredilla are with the Dept. of Signal Theory and Communications, University Carlos III in Madrid, Spain, Email: \{fernando,miguel\}@tsc.uc3m.es.},
Steven Van Vaerenbergh\thanks{Steven Van Vaerenbergh and Ignacio Santamar\'ia are with the Dept. of Communications Engineering, University of Cantabria, Spain, Email: \{steven,nacho\}@gtas.dicom.unican.es.}, Juan Jos\'e Murillo-Fuentes\thanks{Juan Jos\'e Murillo-Fuentes is with the Dept. of Signal Theory and Communications, Spain, Email: murillo@etsi.us.es.}, Miguel L\'azaro-Gredilla and Ignacio Santamar\'ia\thanks{This work has been partially supported by TEC2012-38800-C03-\{01,02\} (ALCIT), TEC2010-19545-C04-03 (COSIMA), TEC2009- 14504-C02-\{01,02\} (DEIPRO), Consolider-Ingenio 2010 CSD2008-00010 (COMONSENS) and MLPM2012 (UE-FP7-PEOPLE-ITN).}}

\markboth{IEEE Signal Processing Magazine, vol. 30, no. 4, July 2013}{}

\maketitle

\IEEEpeerreviewmaketitle

\begin{abstract}
Gaussian processes (GPs) are versatile tools that have been successfully employed to solve nonlinear estimation problems in machine learning, but that are rarely used in signal processing. In this tutorial, we present GPs for regression as a natural nonlinear extension to optimal Wiener filtering. After establishing their basic formulation, we discuss several important aspects and extensions, including recursive and adaptive algorithms for dealing with non-stationarity, low-complexity solutions, non-Gaussian noise models and classification scenarios. Furthermore, we provide a selection of relevant applications to wireless digital communications.
\end{abstract}


\section{Introduction}

Gaussian processes (GPs) are Bayesian state-of-the-art tools for discriminative machine learning, i.e., regression \cite{Williams96}, classification \cite{Kuss05} and dimensionality reduction \cite{Lawrence05}. GPs were first proposed in statistics by Tony O'Hagan \cite{OHagan78} and they are well-known to the geostatistics community as kriging. However, due to their high computational complexity they did not become widely applied tools in machine learning until the early XXI century \cite{rasmusssen2006gaussian}. GPs can be interpreted as a family of kernel methods with the additional advantage of providing a full conditional statistical description for the predicted variable, which can be primarily used to establish confidence intervals and to set hyper-parameters. In a nutshell, Gaussian processes assume that a Gaussian process prior governs the set of possible latent functions (which are unobserved), and the likelihood (of the latent function) and observations shape this prior to produce posterior probabilistic estimates. Consequently, the joint distribution of training and test data is a multidimensional Gaussian and the predicted distribution is estimated by conditioning on the training data.

While GPs are well-established tools in machine learning, they are not as widely used by the signal processing community as neural networks or support vector machines (SVMs) are. In our opinion, there are several explanations for the limited use of GPs in signal processing problems. First, they do not have a simple intuition for classification problems. Second, their direct implementation is computationally demanding. Third, their plain vanilla version might seem uptight and not flexible enough. Fourth, to most signal processing experts Gaussian process merely stands for a noise model and not for a flexible algorithm that they should be using.

In this paper, we present an overview on Gaussian processes explained for and by signal processing practitioners. We introduce GPs as the natural nonlinear Bayesian extension to the linear minimum mean square error (MMSE) and Wiener filtering, which are central to many signal processing algorithms and applications. We believe that GPs provide the correct approach to solve an MMSE filter nonlinearly, because they naturally extend least squares to nonlinear solutions through the kernel trick; they use a simple yet flexible prior to control the nonlinearity; and, evidence sampling or maximization allows setting the hyper-parameters without overfitting. This last feature is most interesting: by avoiding cross-validation we are able to optimize over a larger number of hyperparameters, thus increasing the available kernel expressiveness. Additionally, GP provides a full statistical description of its predictions.

The tutorial is divided in three parts. We have summarized in Figure \ref{fig:diagram_techniques} the relationship between the regression techniques introduced throughout the different sections. In the first part, Section \ref{GP} provides a detailed overview of Gaussian processes for regression (GPR) \cite{Williams96}. We show that they are the natural nonlinear extension to MMSE/Wiener filtering and how they can be solved recursively. The second part of the paper focuses briefly on several key aspects of GP-based techniques. Consecutively, we review solutions to adjust the kernel function (Section \ref{GP_learn} ), to tame the computational complexity of GPs (Section \ref{sparse} ), and to deal with non-Gaussian noise models (Section \ref{WR} ). In the third part, we cover additional extensions of interest to signal processing practitioners, in particular dealing with non-stationary scenarios (Section \ref{tracking} ) and classification problems (Section \ref{gpc} ). We illustrate them with relevant examples in signal processing for wireless communications. We conclude the paper with a discussion.

\begin{figure}[tb]
\centering
\includegraphics[width=8.5cm]{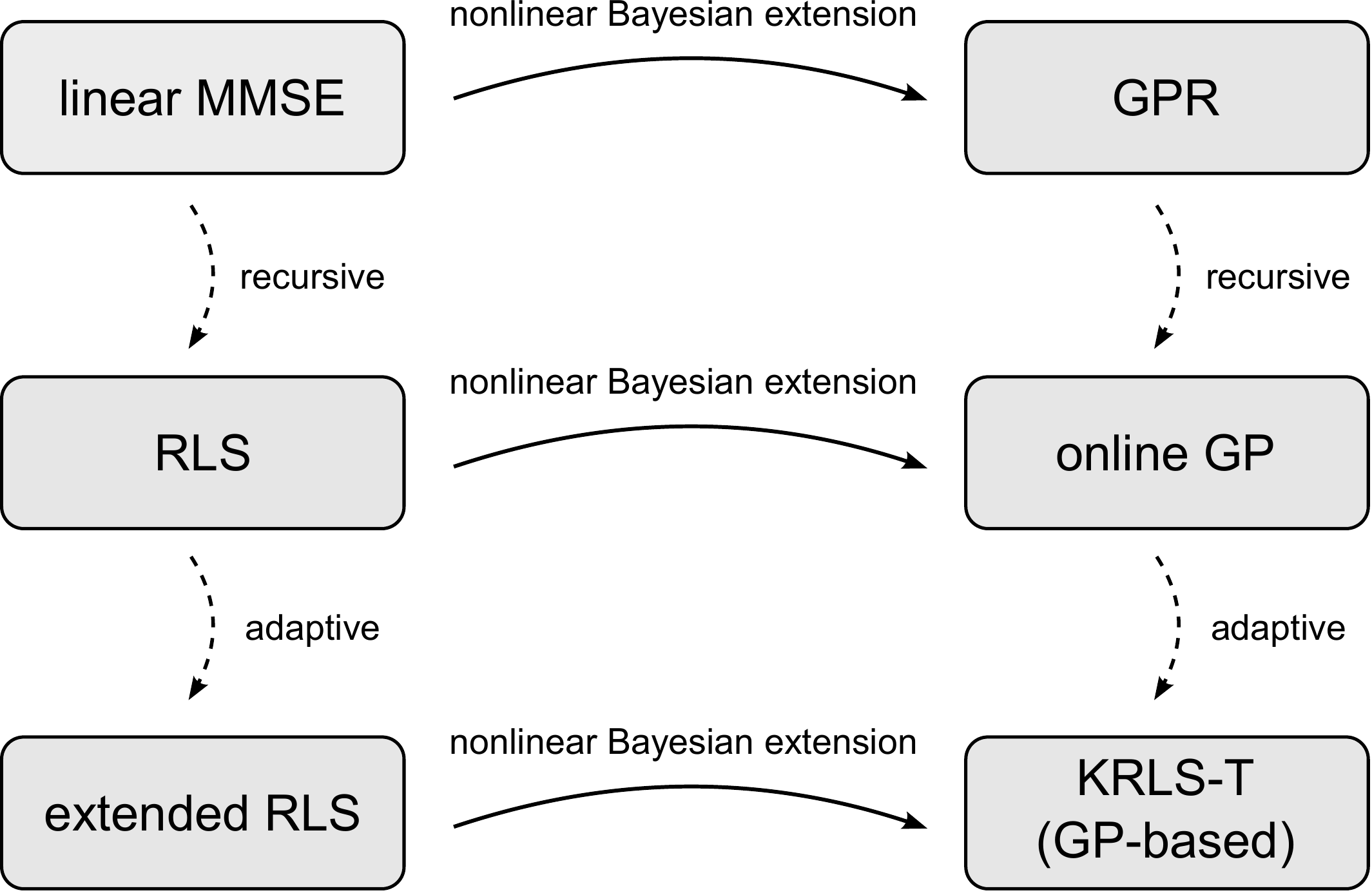}
\caption{Relationship between the regression techniques discussed in this tutorial.}
\LABFIG{diagram_techniques}
\end{figure}



\section{Gaussian Processes for Machine Learning}\label{GP}

\subsection{Minimum mean square error: a starting point}

GPs can be introduced in a number of ways and we, as signal processing practitioners, find it particularly appealing to start from the MMSE solution. This is because the Wiener solution, which is obtained by minimizing the MSE criterion, is our first approach to most estimation problems and, as we show, GPs are its natural Bayesian extension.

Many signal processing problems reduce to estimating from an observed random process $\x \in \mathbb{R}^{p}$ another related process $\y \in \mathbb{R}$. These two processes are related by a probabilistic, possibly unknown,
model $p(\x|\y)$. It is well known that the unconstrained MMSE estimate,
\begin{align}\label{mmse}
\arg\!\min_{f(\x)}E\left[\left\|\y-f(\x)\right\|^2\right],
\end{align}
coincides with the conditional mean estimate of $\y$ given $\x$
\begin{equation}\label{con_mean}
f_{mmse}(\x)=E[\y|\x]=\int \y p(\y|\x) d\y=\int \y \frac{p(\x|\y)p(\y)}{p(\x)} d\y.
\end{equation}
If $p(\y,\x)$ is jointly Gaussian, i.e. $p(\y)$ and $p(\x|\y)$ are Gaussians and $E[\x|\y]$ is linear in $\y$, this solution is linear. If $\y$ and $\x$ are zero mean, the solution yields $E[\y|\x]=\w^\top \x$, where
\begin{align}
\w_{mmse}&=\arg\!\min_{\w}E\left[\left(\y-\w^\top\x\right)^2\right]
=\left(E\left[\x\x^\top\right]\right)^{-1} E\left[\x \y\right].
\label{mmse_lin}
\end{align}
Furthermore, these expectations can be easily estimated, using the sample mean, from independently and identically distributed (iid) samples drawn from $p(\x|\y)$ and $p(\y)$, namely $\tset_n=\{\x_{\indexData}, \y_{\indexData}\}^{n}_{\indexData=1}$.

However, if $\x$ is not linearly related to $\y$ (plus Gaussian noise) or $\y$ is not Gaussian distributed, the conditional estimate of $\y$ given $\x$ is no longer linear. Computing the nonlinear conditional mean estimate in \eqref{con_mean} directly from $\tset_n$ either leads to overfitted solutions, because there are no convergence guarantees for general density estimation \cite{Vapnik98}, or to suboptimal solutions, if we restrict the density model to come from a narrow class of distributions. For instance, in channel equalization, although suboptimal, the sampled version of \eqref{mmse_lin} is used due to its simplicity. One viable solution would be to minimize the sampled version of \eqref{mmse} with a restricted family of approximating functions to avoid overfitting. Kernel least squares (KLS) \cite{PerezCruz04b} and Gaussian process regression, among others, follow such approach.

\subsection{Gaussian Processes for Regression}\label{gpr_sec}

In its simplest form, GPR models the output nonlinearly according to
\begin{equation}\LABEQ{GLR}
\y=f(\x)+\noiseOut ,
\end{equation}
and it follows \eqref{mmse}, without assuming that $\x$ and $\y$ are linearly related or that $p(\y)$ is Gaussian distributed. Nevertheless, it still considers that $p(\y|\x)$ is Gaussian distributed, i.e., $\noiseOut$ is a zero-mean Gaussian\footnote{A further relaxation to this condition is discussed in Section \ref{WR}.}. In this way, GP can be understood as a natural nonlinear extension to MMSE estimation. Additionally, GPR does not only estimate \eqref{con_mean} from $\tset_n$, but it also provides a full statistical description of $\y$ given $\x$, namely
\begin{equation}\LABEQ{1r}
p(\y|\x,\tset_n).
\end{equation}



GPs can be presented as a nonlinear regressor that expresses the input-output relation in \EQ{GLR} by assuming that a real-valued function $f(\x)$, known as \emph{latent
function}, underlies the regression problem and that this function
follows a Gaussian process. Before the labels are revealed, we
assume this latent function has been drawn from a Gaussian
process prior. GPs are characterized by their mean and covariance functions, denoted by $\mu(\x)$ and $\kernel(\x,\x')$, respectively. Even though nonzero mean priors might be of use, working with zero-mean priors typically represents a reasonable assumption and it simplifies the notation.
The covariance function explains the correlation between each pair of
points in the input space and characterizes the functions that can be described by the
Gaussian process.  For example, $\kernel(\x,\x')=\x^\top\x'$ only yields
linear latent functions and it is used to solve Bayesian linear
regression problems, for which the mean of the posterior process coincides with the MMSE solution in \eqref{mmse_lin}, as shown in Section \ref{wsv}. We cover the design of covariance functions in Section \ref{GP_learn}.


For any finite set of inputs $\tset_n$, a Gaussian process becomes a
multidimensional Gaussian defined by its mean (zero in our case) and
covariance matrix, $(\K_n)_{ij}=\kernel(\x_i,\x_j),\forall\x_i,\x_j\in\tset_n$. The Gaussian process prior becomes
\begin{equation}\LABEQ{GP_prior}
p(\fp_n|\X_n)=\mathcal{N}(\mathbf{0},\K_n),
\end{equation}
where $\fp_n=[\fv(\x_1), \fv(\x_2),\ldots, \fv(\x_n)]^\top$ and $\X_n=[\x_1, \x_2, \ldots, \x_n]$. We want to compute the estimate for a general input $\x$, when the labels for the $n$ training examples, denoted by $\yv_n=[\y_1, \y_2, \ldots, \y_n]^\top$, are known.
We can analytically compute \EQ{1r} by using the standard tools of Bayesian statistics:
Bayes' rule, marginalization and conditioning.

We first apply Bayes' rule to obtain the posterior density for the latent function
\begin{equation}\LABEQ{posteriorf}
p(\fv(\x),\fp_n |  \x,\tset_n) = \frac{ p(\yv_n| \fp_n) p( \fv(\x),\fp_n|  \x,\X_n)}{p(\yv_n|\X_n)},
\end{equation}
where $p( \fv(\x),\fp_n| \x,\X_n)$ is the Gaussian process prior in \EQ{GP_prior} extended with a general input $\x$, $p(\yv_n| \fp_n)$ is the likelihood for the latent function at the training set, in which $\yv_n$ is independent of $\X_n$ given the latent function $\fp_n$, and $p(\yv_n|\X_n)$ is the marginal likelihood or evidence of the model.

The likelihood function is given by a factorized model:
\begin{equation}\LABEQ{likef}
p(\yv_n|\fp_n)=\prod_{\indexData=1}^n p(\y_\indexData|\fv(\x_\indexData)),
\end{equation}
because the samples in $\tset_n$ are iid. In turn, for each pair $(f(\x_i),\y_i)$, the likelihood is given by \EQ{GLR}, therefore
\begin{equation}\LABEQ{like_ind}
p(\y_\indexData|\fv(\x_\indexData))\sim\mathcal{N}(\fv(\x_\indexData),\sigma_\noiseOut^2).
\end{equation}
A Gaussian likelihood function is conjugate to the Gaussian prior and hence the posterior in \EQ{posteriorf} is also a multidimensional Gaussian, which simplifies the computations to obtain \EQ{1r}. If the observation model were not Gaussian, warped Gaussian processes (see Section \ref{WR} ) could be used to estimate \EQ{1r}.

Finally, we can obtain the posterior density in \EQ{1r} for a general input $\x$ by conditioning on the training set and $\x$, and by marginalizing
the latent function:
\begin{eqnarray}\LABEQ{6r}
p(\y|\x,\tset_n)\!\!
=\!\!\int p(\y|\fv(\x)) p(\fv(\x)|\x,\tset_n)d\fv(\x),
\end{eqnarray}
where\footnote{Given the training data set, ${\bf{f}}_n$ takes values in $\mathbb{R}^{n}$ as it is a vector of $n$ samples of a Gaussian process.}
\begin{eqnarray}\LABEQ{5r}
p(\fv(\x)|\tset_n,\x)=\int p(\fv(\x), \fp_n| \x,\tset_n) d\fp_n.
\end{eqnarray}

We have divided the marginalization in two separate equations to show the
marginalization of the latent function over the training set in
\EQ{5r}, and the marginalization of the latent function at a general input $\x$ in \EQ{6r}. As mentioned earlier, the likelihood and the prior
are Gaussians and therefore the marginalization in \EQ{6r} and
\EQ{5r} only involves Gaussian distributions. Thereby, we can
analytically compute \EQ{6r} and \EQ{5r} by using Gaussian conditioning and
marginalization properties, leading to the following Gaussian density for the output:
\begin{equation}\LABEQ{6r2}
p(\fv(\x)|\x,\tset_n)\sim\mathcal{N}\left(\mu_{\fv(\x)},\sigma_{\fv(\x)}^2\right),
\end{equation}
where
\begin{subequations}
\begin{align}
\mu_{\fv(\x)}&=\kk^\top\mathbf{C}_n^{-1}\yv_n, \LABEQ{pmean1} \\
\sigma_{\fv(\x)}^2&=k(\x,\x)- \kk^\top\mathbf{C}_n^{-1}\kk,\LABEQ{pcov1}
\end{align}
\end{subequations}
with
\begin{align}\label{kk}
\kk&=[k(\x_1,\x), k(\x_2,\x), \ldots, k(\x_n,\x)]^\top,\\
\mathbf{C}_n&=\K_n+\sigma_\noiseOut^2\mathbf{I}_n.
\end{align}
The mean for $p(\y|\x,\tset_n)$ is also given by \EQ{pmean1}, i.e., $\mu_y=\mu_{\fv(\x)}$, and its variance is
\begin{equation}\label{pcov2}
\sigma_{\y}^2=\sigma_{\fv(\x)}^2+\sigma_\nu^2,
\end{equation}
which, as expected, also accounts for the noise in the observation model.

The mean prediction of GPR in \EQ{pmean1} is the solution provided by KLS, or kernel ridge regression (KRR) \cite{PerezCruz04b}, in which the covariance function takes the place of the kernel. However, unlike standard kernel methods, GPR provides error bars for each estimate in \EQ{pcov1} or \eqref{pcov2} and has a natural procedure for setting the covariance/kernel by evidence sampling or maximization, as detailed in Section \ref{GP_learn}. In SVM or KRR the hyper-parameters are typically adjusted by cross-validation, needing to retrain the models for different settings of the hyper-parameters on a grid search. So, typically only one or two hyper-parameters can be fitted. GPs can actually learn tens of hyper-parameter, because either sampling or evidence maximization allows setting them by a hassle-free procedure.

\subsection{An example}

In \FIG{gpr} we include an illustrative example with 20 training points, in which  we depict \EQ{6r2} for any $\x$ between $-3$ and $4$. We used standard functions from the GPML toolbox, available at \url{http://www.gaussianprocess.org/gpml/}, to generate the GP in this figure. We have chosen a Gaussian kernel that is fixed\footnote{The kernel is typically expressed in a parametric form, see Section \ref{GP_learn}.} as $\kernel(\x_i,\x_j)= \exp{(-2||\x_i-\x_j||^2)}$ and $\sigma_\noiseOut=0.1$. In the plot, we show the mean of the process in red and the shaded area denotes the error bar for each prediction, i.e., $\mu_{\y}\pm2\sigma_{\y}$. We also plot 5 samples from the posterior in thin blue lines.

We observe three different regions in the figure. On the right-hand side, we do not have samples and, for $\x>3$, the GPR provides the solution given by the prior  (zero mean and $\pm2$). At the center, where most of the data points lie, we have a very accurate view of the latent function with small error bars (close to $\pm2\sigma_\noiseOut$). On the left hand side, we only have two samples and we notice the mixed effect of the prior widening the error bars and the data points constraining the values of the mean to lie close to the available samples. This is the typical behavior of GPR, which provides an accurate solution where the data lies and high error bars where we do not have available information and, consequently, we presume that the prediction in that area is not accurate.

\begin{figure}[tb]
\centering
\includegraphics[width=8.5cm]{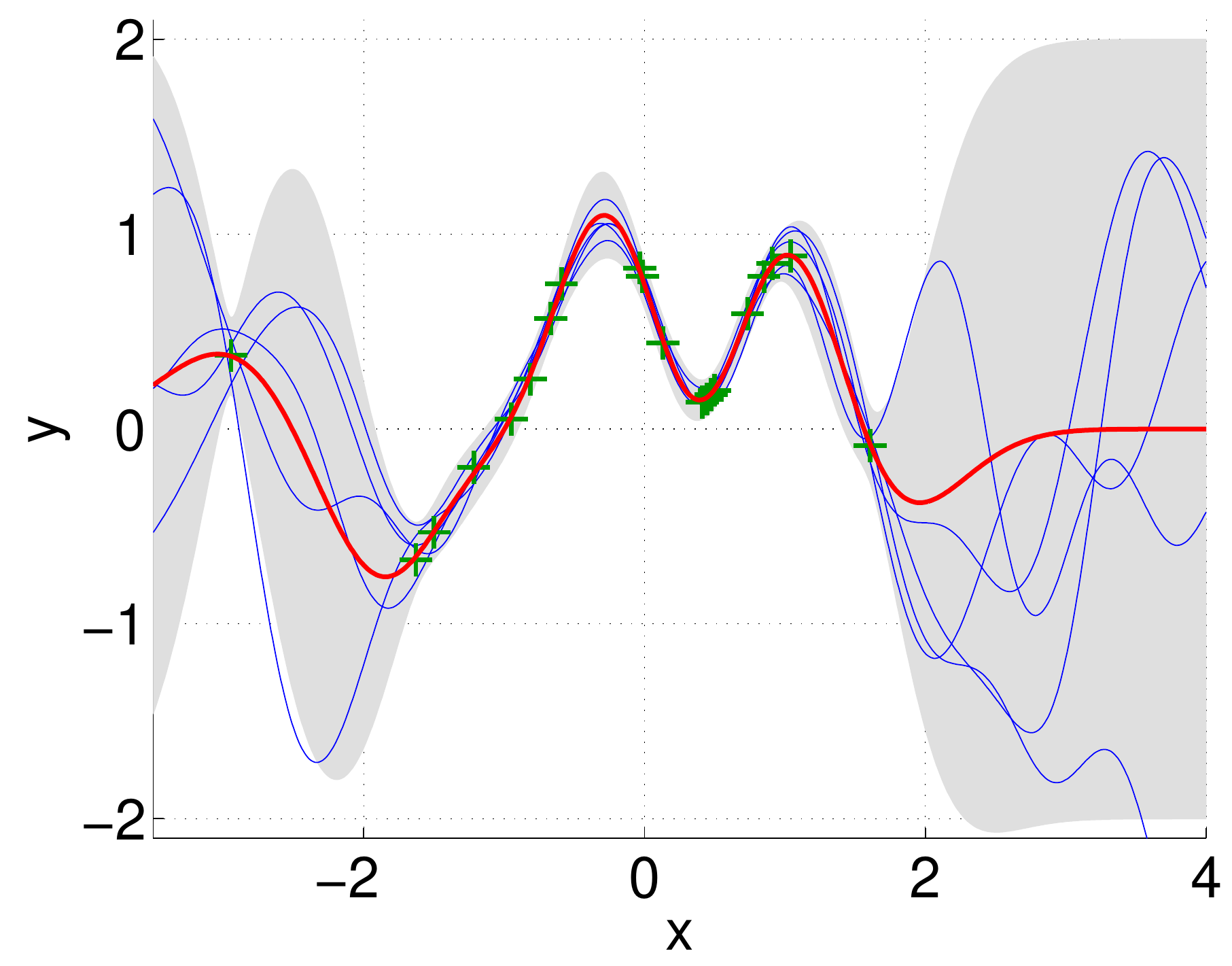}
\caption{Example of a Gaussian process posterior in \EQ{6r2} with 20 training samples, denoted by green $+$. Five instances of the posterior are plotted by thin blue lines and the mean of the posterior, $\mu_{\y}$, by a red thick line. The shaded area denotes the error bars for the mean prediction: $\mu_{\y}\pm2\sigma_{\y}$. } \LABFIG{gpr}
\end{figure}

\subsection{Recursive GPs} \label{sec:adaptive}

In many signal processing applications, the samples become available sequentially and estimation algorithms should obtain the new solution every time a new datum is received. In order to keep the computational complexity low, it is more interesting to perform inexpensive recursive updates rather than to recalculate the entire batch solution. {\em Online Gaussian Processes} \cite{csato2001sparse} fulfill these requisites as follows. 

Let us assume that we have observed the first $n$ samples and that at this point the new datum $\x_{n+1}$ is provided. We can readily compute the predicted distribution for $\y_{n+1}$ using \EQ{pmean1}, \EQ{pcov1} and \eqref{pcov2}. Furthermore, by using the formula for the inverse of a partitioned matrix and the Woodbury identity we update $\mathbf{C}_{n+1}^{-1}$ from $\mathbf{C}_{n}^{-1}$
\begin{equation}\label{update_C}
\mathbf{C}_{n+1}^{-1}=\mtt{ \mathbf{C}_n^{-1}+\mathbf{C}_n^{-1}\kk_{n+1}\kk_{n+1}^\top\mathbf{C}_n^{-1}/\sigma_{y_{n+1}}^2 }{-\mathbf{C}_n^{-1}\kk_{n+1}/\sigma_{y_{n+1}}^2}{-\kk_{n+1}^\top\mathbf{C}_n^{-1}/\sigma_{y_{n+1}}^2}{1/\sigma_{y_{n+1}}^2},
\end{equation}
where $\sigma_{y_{n+1}}^2$ and $\kk_{n+1}$ correspond to \eqref{pcov2} and \eqref{kk}, respectively, for $\x = \x_{n+1}$.

Nevertheless, for online scenarios, it is more convenient to update the predicted mean and covariance matrix for all the available samples, as it is easier to interpret how the prediction changes with each new datum. Additionally, as we will show in Section \ref{tracking}, this formulation makes the adaptation to non-stationary scenarios straightforward. Let us denote by $\pmean_n$ and $\pcov_n$ the posterior mean and covariance matrix for the samples in $\tset_n$. By applying \EQ{pmean1} and \EQ{pcov1} we obtain
\begin{subequations}
\label{eq:post}
\begin{align}
\pmean_n&=\K_n\mathbf{C}_{n}^{-1}\yv_n,\\
\pcov_n&=\K_n-\K_n\mathbf{C}_{n}^{-1}\K_n, \label{eq:postsigma}
\end{align}
\end{subequations}
Once the new datum $(\x_{n+1}, y_{n+1})$ is observed, the updated mean and covariance matrix can be computed recursively as follows:
\begin{subequations}
\label{eq:postupdate}
\begin{align}
\pmean_{n+1}&=\vt{\pmean_n}{\mu_{f(\x_{n+1})}} - \frac{\mu_{f(\x_{n+1})} - y_{n+1}}{\sigma_{y_{n+1}}^2} \vt{\h_{n+1}}{\sigma_{f(\x_{n+1})}^2} \label{eq:meanupdate}\\
\pcov_{n+1}&= \mtt{\pcov_n}{\h_{n+1}}{\h_{n+1}^\top}{\sigma_{f(\x_{n+1})}^2} - \frac{1}{\sigma_{y_{n+1}}^2}\vt{\h_{n+1}}{\sigma_{f(\x_{n+1})}^2} \left[ \h_{n+1}^\top \ \ \sigma_{f(\x_{n+1})}^2\right], \label{eq:covupdate}
\end{align}
\end{subequations}
where $\h_{n+1}=\pcov_{n}\K_n^{-1}\kk_{n+1}=(\mathbf{I}_n-\K_n\mathbf{C}_n^{-1})\kk_{n+1}$. As can be observed in \eqref{eq:meanupdate}, the mean of the new process is obtained by applying a correction term to the previous mean, proportional to the estimation error, $\mu_{f(\x_{n+1})} - y_{n+1}$. Because of the relation between $\pcov_n$ and $\mathbf{C}_{n}^{-1}$ stated in \eqref{eq:postsigma}, only one of the two matrices needs to be stored and updated in an online formulation. Some authors \cite{csato2001sparse} prefer to rely on $\mathbf{C}_{n}^{-1}$, whereas others \cite{vanvaerenbergh2012kernel} store and update $\pcov_n$.

The recursive update of the mean in \eqref{eq:meanupdate} is equivalent to what is known as {\em kernel recursive least-squares} (KRLS) in the signal processing literature (see for instance \cite{csato2001sparse,engel2004kernel,vanvaerenbergh2012kernel} ). The unbounded growth of the involved matrices, visible in \eqref{eq:postupdate} and \eqref{update_C}, is the main limitation in the KRLS formulation. Practical KRLS implementations typically either limit this growth \cite{engel2004kernel,liu2009information} or even fix the matrix sizes \cite{vanvaerenbergh2006sliding}. Nevertheless, the solution of KRLS is limited to the mean only and it cannot estimate confidence intervals. By using a GP framework, though, an estimate of the entire posterior distribution is obtained, including the covariance in \eqref{eq:covupdate}.

\subsection{Connection to MMSE: GPR with a linear latent function}\label{wsv}

If we replace $f(\x)$ in \EQ{GLR} with a linear model
\begin{equation*}\label{GLRM}
\y=\w^\top\x+\noiseOut,
\end{equation*}
the Gaussian process prior over $f(\x)$ becomes a spherical-Gaussian prior distribution over $\w$, $p(\w) \sim \mathcal{N} (\mathbf{0}, \sigma_{\w}^2\id)$.

We can now compute the posterior for $\w$, as we did for the latent function in \EQ{posteriorf}
\begin{align*}
p(\w|\mathcal{D}) = \frac{p(\yv|\X,\w)p(\w)}{p(\yv|\X)} 
= \frac{p(\w)}{p(\yv|\X)} \prod_{i=1}^n p(\y_i|\x_i,\w),
\end{align*}
where $p(\y_i|\x_i,\w)$ is the likelihood. 
Since the prior and likelihood are Gaussians, so it is the posterior, and its mean and covariance are given by
\begin{subequations}
\begin{align}
\muw &=\frac{1}{\sigma_\noiseOut^2} {\mathbf \Sigma}_\w \X^\top\yv, \label{med_w}\\
{\mathbf \Sigma}_\w &= \left(\X^\top\X/\sigma_\noiseOut^2 + \id/\sigma_\w^2\right)^{-1} \label{sigma_w}.
\end{align}
\end{subequations}
We can readily notice that \eqref{med_w} is the sampled version of \eqref{mmse_lin}, when the prior variance $\sigma_\w^2$ tends to infinity (i.e., the prior has no effect of the solution). The precision matrix (the inverse covariance) is composed of two terms: the first depends on the data and the other one on the prior over $\w$. The effect of the prior in the mean and covariance fades away, as we have more available data. The estimate for a general input $\x$ is computed as in \EQ{6r}
\begin{eqnarray}\LABEQ{6r_lin}
p(\y|\x,\tset)\!\!
=\!\!\int p(\y|\x,\w) p(\w|\tset)d\w,
\end{eqnarray}
which is a Gaussian distribution with mean and variance given by:
\begin{align}
\mu_y&= \x^\top\muw \label{mu_y_wei}=\frac{1}{\sigma_\noiseOut^2} \x^\top {\mathbf \Sigma}_\w \X^\top\yv \\
\sigma_y^2&=\x^\top{\mathbf \Sigma}_\w\x+\sigma_\noiseOut^2\label{sig_y_wei}
\end{align}
Equations \eqref{mu_y_wei} and \eqref{sig_y_wei} can be, respectively, rewritten as \EQ{pmean1} and \eqref{pcov2}, if we use the inner product between the $\x_i$ multiplied by the width of the prior over $\w$, i.e. the kernel matrix is given by: $\K_n=\X\sigma_\w^2\id\X^\top$. The kernel matrix must include the width of the prior over $\w$, because the kernel matrix represents the prior of the Gaussian process and $\sigma_\w^2$ is the prior of the linear Bayesian estimator. By using the Woodbury's identity, it follows that
  \begin{align}
{\mathbf \Sigma}_\w &= \sigma_\w^2\id-\sigma_\w^2\id\X^\top \left(\sigma_\noiseOut^2\id+\K_n\right)^{-1}\X\sigma_\w^2\id \label{sigma_w1}.
\end{align}
Now, by replacing \eqref{sigma_w1} in \eqref{mu_y_wei} and \eqref{sig_y_wei}, we,respectively, recover \EQ{pmean1} and \eqref{pcov2}. These steps connect the estimation of a Bayesian linear model and the nonlinear estimation using a kernel or covariance function without needing to explicitly indicate the nonlinear mapping.

\section{Covariance functions}\label{GP_learn}

In the previous section, we have assumed that the covariance functions $\kernel(\x,\x')$ are known, which is not typically the case. In fact, the design of a good covariance function is crucial for GPs to provide accurate nonlinear solutions. The covariance function plays the same role as the kernel function in SVMs or KLS \cite{PerezCruz04b}. It describes the relation between the inputs and its form determines the possible solutions of the GPR. It controls how fast the function can change or how the samples in one part of the input space affect the latent function everywhere else. For most problems, we can specify a parametric kernel function that captures any available information about the problem at hand. As already discussed, unlike kernel methods, GPs can infer these parameters, the so-called \emph{hyper-parameters}, from the samples in $\tset_n$ using the Bayesian framework. Instead of relying on computational intensive procedures as cross-validation \cite{Kimeldorf71} or learning the kernel matrix \cite{Bousquet03}, as kernel methods need to.

The covariance function must be positive semi-definite, as it represents the covariance matrix of a multidimensional Gaussian distribution. The covariance can be built by adding simpler covariance matrices, weighted by a positive hyper-parameter, or by multiplying them together, as the addition and multiplication of positive definite matrices yields a positive definite matrix. In general, the design of the kernel should rely on the information that we have for each estimation problem and should be designed to get the most accurate solution with the least amount of samples. Nevertheless, the following kernel in \EQ{15} often works well in signal processing applications
\begin{equation}\LABEQ{15}
\kernel(\x_i,\x_j)=\alpha_{1}\exp\left(-\sum_{\ell=1}^{d}\gamma_{\ell}
||x_{i\ell}-x_{j\ell}||^{2}\right)+\alpha_{2}\x_i^{\top}\mathbf{x}_j+\alpha_3\delta_{ij},
\end{equation}
where
$\boldsymbol{\theta}=[\alpha_{1},\gamma_{1},\gamma_{2},\ldots,\gamma_{d},\alpha_{2},\alpha_3]^\top$ are the hyper-parameters. The first term is a radial basis kernel, also denoted as RBF or Gaussian, with a different length-scale for each input dimension. This term is universal and allows constructing a generic nonlinear regressor. If we have symmetries in our problem, we can use the same length-scale for all dimensions: $\gamma_{\ell}=\gamma$ for $\ell =1, \ldots, d$. The second term is the linear covariance function. The last term represents the noise variance $\alpha_3 = \sigma^2_\nu$, which can be treated as an additional hyper-parameter to be learned from the data. We can add other terms or other covariance functions that allow for faster transitions, like the Mat\'ern kernel among others \cite{rasmusssen2006gaussian}.

If the hyper-parameters, $\boldsymbol{\theta}$, are unknown, the likelihood in \EQ{likef} and the prior in \EQ{GP_prior} can, respectively, be expressed\footnote{We have dropped the subindex $n$, as it is inconsequential and unnecessarily clutters the notation.} as $p(\yv|\fp,\boldsymbol{\theta})$ and $p(\fp| \X,\boldsymbol{\theta})$, and we can proceed to integrate out $\boldsymbol{\theta}$ as we did for the latent function, $\fp$, in Section \ref{gpr_sec}. First, we compute the \emph{marginal likelihood} of the hyper-parameters of the kernel given the training dataset
\begin{align}\LABEQ{like_t}
p(\yv|\X,\boldsymbol{\theta})=\int& p(\yv| \fp, \boldsymbol{\theta}) p(\fp| \X,\boldsymbol{\theta}) d\fp.
\end{align}
Second, we can define a prior for the hyper-parameters, $p(\boldsymbol{\theta})$, that can be used to construct its posterior. Third, we integrate out the hyper-parameters to obtain the predictions.  However, in this case, the marginal likelihood does not have a conjugate prior and the posterior cannot be obtained in closed form. Hence, the integration has to be done either by sampling or approximations. Although this approach is well principled, it is computational intensive and it may be not feasible for some applications. For example, Markov-Chain Monte Carlo (MCMC) methods require several hundred to several thousand samples from the posterior of $\boldsymbol{\theta}$ to integrate it out.
Interested readers can find further details in \cite{rasmusssen2006gaussian}.

Alternatively, we can maximize the marginal likelihood in \EQ{like_t}  to obtain its optimal setting \cite{Williams96}. Although setting the hyper-parameters by maximum likelihood (ML) is not a purely Bayesian solution, it is fairly standard in the community and it allows using Bayesian solutions in time sensitive applications. This optimization is nonconvex \cite{MacKay03}, but, as we increase the number of training samples, the likelihood becomes a unimodal distribution around the ML hyper-parameters and the solution can be found using gradient ascent techniques. See \cite{rasmusssen2006gaussian} for further details.


\section{Sparse GPs: Dealing with large-scale data sets}\label{sparse}

To perform inference under any GP model, the inverse of the covariance matrix must be computed. This is a costly operation, $\bigO(n^3)$, that becomes prohibitive for large enough $n$. Given the ever-increasing availability of large-scale databases, a lot of effort has been devoted over the last decade to the development of approximate methods that allow  inference in GPs  to scale linearly with the number of data points. These approximate methods are  referred to as ``sparse GPs'', since they approximate the full GP model using a finite-basis-set expansion. This set of bases is usually spawned by using a common functional form with different parametrizations. For instance, it is common to use bases of the type $\{k(\mathbf{z}_b, \x)\}_{b=1}^m$, where $\{\mathbf{z}_b\}_{b=1}^m$ ---known as the \emph{active set}--- is a subset of the input samples parametrizing the bases.

Under the unifying framework of \cite{Candela05}, it can be shown that most relevant sparse GP proposals \cite{
 Seeger03, Snelson06},  which  were initially thought of as entirely different low-cost approximations, can be expressed as exact inference under different modifications of the original GP prior. This modified prior  induces a rank-$m$ ($m\ll n$) covariance matrix  ---plus optional (block) diagonal correcting terms---, clarifying how the  reduced $\bigO(m^2n)$ cost of exact inference arises.

Among the mentioned approximations, the sparse pseudo-input GP (SPGP) \cite{Snelson06} is generally regarded as the most efficient. Unlike other alternatives, it does not require the active set to be a subset of the training data. Instead, $\{\mathbf{z}_b\}_{b=1}^m$ can be selected to lie anywhere in the input space, thus increasing the flexibility of the finite set expansion. This selection is typically performed by evidence maximization. An even more flexible option, which does not require the active set to even lie in the input domain, is presented in \cite{Lazaro09}.

Despite the success of SPGP, it is worth mentioning that increasing the number of bases in this algorithm does not yield, in general,  convergence to the full GP solution  because the active set $\{\mathbf{z}_b\}_{b=1}^m$ is not constrained to be a subset of input data. This might lead to overfiting in some pathological cases. A recent variational sparse GP proposal  that guarantees convergence to the full GP solution  while still allowing the active set to be unconstrained is presented in \cite{Titsias09}.

Further approaches yielding reduced computational cost involve numerical approximations to accelerate matrix-vector multiplications and compactly supported covariance functions which set most entries of  the covariance matrix to zero \cite{Gu12}.

Sparsity is often seen in online signal processing in the form of \emph{pruning}, which restricts the active set to  a subset of input data. The success of SPGP and its variational counterpart suggests that  advanced forms of pruning may result in increased efficiency for a given sparsity level.

\section{Warped GPs: Beyond the standard noise model} \label{WR}

Even though GPs are very flexible priors for the latent function, they might not be suitable to model all types of data. It is often the case that applying a logarithmic transformation to the target variable of some regression task (e.g., those involving stock prices, measured sound power, etc) can enhance the ability of GPs to model it.

In \cite{Snelson03} it is shown that it is possible to include a non-linear preprocessing of output data $h(y)$ (called \emph{warping function} in this context)  as part of the modeling process and learn it. In more detail, a parametric form for $z = h(y)$ is selected, then $z$ (which depends on the parameters of $h(y)$) is regarded as a GP, and finally, the parameters of $h(y)$ are selected by maximizing the evidence of such GP (i.e., a ML approach). The authors suggest using
$
h(y) = \sum_{i=1}^l a_i \tanh (b_i(y+c_i))
$
as the parametric form of the warping function, but any option resulting in a monotonic function is valid. A non-parametric version of warped GPs using a variational approximation has been  proposed in \cite{Lazaro12}.

\section{Tracking non-stationary scenarios: Learning to forget}\label{tracking}

KRLS algorithms, discussed in Section~\ref{sec:adaptive}, traditionally consider that the mapping function $f(\cdot)$ is constant throughout the whole learning process \cite{engel2004kernel,liu2010adaptive}. However, in the signal processing domain this function (which might represent, for instance, a fading channel) is often subject to changes and the model must account for this non-stationarity. Some kernel-based algorithms have been proposed to deal with non-stationary scenarios. They include a kernelized version of the extended RLS filter \cite{liu2010adaptive}, a sliding-window KRLS approach \cite{vanvaerenbergh2006sliding} and a family of projection-based algorithms \cite{slavakis2008online,theodoridis2011adaptive}.

In order to add adaptivity to the online GP algorithm described in Section \ref{sec:adaptive}, it is necessary to make it ``forget'' the information contained in old samples. This becomes possible by including a ``forgetting'' step after each update
\begin{subequations}
\label{eq:bttpforgetting}
\begin{align}
\pmean &\leftarrow \sqrt{\lambda}\pmean\\
\pcov &\leftarrow \lambda\pcov+(1-\lambda)\Kb.
\end{align}
\end{subequations}
to shift the posterior distribution towards the prior (for $0<\lambda<1$), thus effectively reducing the influence of older samples. Note that when using this formulation there is no need to store or update $\mathbf{C}^{-1}$, see \cite{vanvaerenbergh2012kernel} for further details. The adaptive, GP-based algorithm obtained in this manner is known as KRLS-T. 

Equations \eqref{eq:bttpforgetting} might seem like an {\em ad-hoc} step to enable forgetting. However, it can be shown that the whole learning procedure ---including the mentioned controlled forgetting step--- corresponds exactly to a principled non-stationary scheme within the GP framework, as described in \cite{vanvaerenbergh2012estimation}. It is sufficient to consider an augmented input space that includes the time stamp $t$ of each sample and define a \emph{spatio-temporal} covariance function:
\begin{equation}
\label{eq:stcov}
k_\text{st}([t~~\x^\top]^\top, [t' ~~\x'^\top]^\top) = k_\text{t}(t,t') k_\text{s}(\x,\x'),
\end{equation}
where $k_\text{s}(\x,\x')$ is the already-known spatial covariance function and $k_\text{t}(t,t')$ is a temporal covariance function giving more weight to samples that are closer in time. Inference on this augmented model effectively accounts for non-stationarity in $f(\cdot)$ and recent samples have more impact in predictions for the current time instant. It is fairly simple to include this augmented model in the online learning process described in the previous section. When the temporal covariance is set to $k_\text{t}(t,t') = \lambda^\frac{|t-t'|}{2},~~\lambda \in (0,1]$, inference in the augmented spatio-temporal GP model is exactly equivalent to using \eqref{eq:bttpforgetting} after each update \eqref{eq:postupdate} in the algorithm of Section \ref{sec:adaptive}, which has the added benefit of being inexpensive and online. See \cite{lazaro2011bayesian,vanvaerenbergh2012kernel,vanvaerenbergh2012estimation} for further details.

Observe that $\lambda$ is used here to model the speed at which $f(\cdot)$ varies, playing a similar r\^ole to that of the forgetting factor in linear adaptive filtering algorithms. When used with a linear spatial covariance, the above model reduces to linear extended RLS filtering. The selection of this parameter is usually rather {\em ad-hoc}. However, using the GP framework, we can select it in a principled manner using Type-II ML, see \cite{vanvaerenbergh2012estimation}.

In \FIG{b2p} we take the example of \FIG{gpr} and we apply a forgetting factor $\lambda=0.8$. The red continuous line indicates the original mean function before forgetting. After applying one forgetting update, this mean function is displaced toward zero, as indicated by the the blue dashed line. The shaded gray area represents the error bars prior to forgetting. The forgetting update expands this area into the shaded red area, which tends to the prior variance of 1.

\begin{figure}[tb]
\centering
\includegraphics[width=8cm]{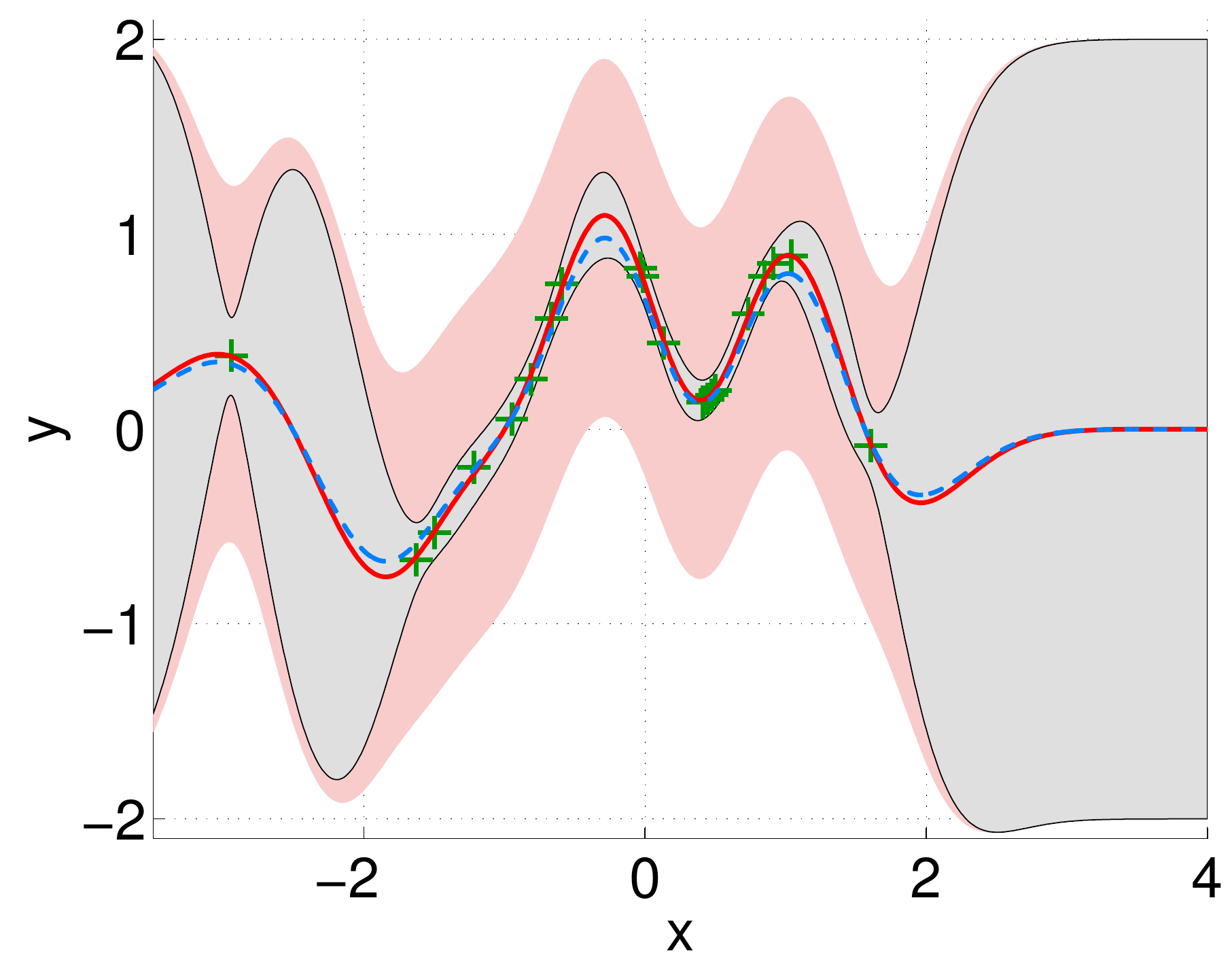}
\caption{Illustration of forgetting step \eqref{eq:bttpforgetting} on the GP of Fig.~\ref{fig:gpr}: the dashed line represents the predictive mean that is pulled towards the prior mean, while the shaded red area represents the region $\mu_{\y}\pm2\sigma_{\y}$ after forgetting.} \LABFIG{b2p}
\end{figure}

\subsection{Tracking a time-selective nonlinear communication channel}

To illustrate the validity of the adaptive filtering algorithm, we focus on the problem of tracking a nonlinear Rayleigh fading channel \cite[Chapter 7]{sayed2003fundamentals}.
The used model consists of a memoryless saturating nonlinearity followed by a time-varying linear channel, as shown in Fig.~\ref{fig:diagram_rayleigh}.
This model appears for instance in broadcast or satellite communications when the amplifier operates close to saturation regime \cite{feher83digital}.


In a first, simulated setup, the time-varying linear fading channel consists of $5$ randomly generated paths, and the saturating nonlinearity is chosen as $y=\tanh(x)$. We fix the symbol rate at $T = 1\mu$s, and we simulate two scenarios: one with a normalized Doppler frequency of $f_dT = 10^{-4}$ (where $f_d$ denotes the Doppler spread), representing a slow-fading channel, and another one with $f_dT=10^{-3}$, corresponding to a fast time-varying channel. Note that a higher Doppler frequency yields a more difficult tracking problem, as it corresponds to a channel that changes faster in time. We consider a Gaussian source signal, and we add $30$ dB of additive white Gaussian noise to the output signal. Given one input-output data pair per time instant, the tracking problem consists in estimating the received signal that corresponds to a new channel input. 

\begin{figure}[tb]
\centering
\includegraphics[width=8cm]{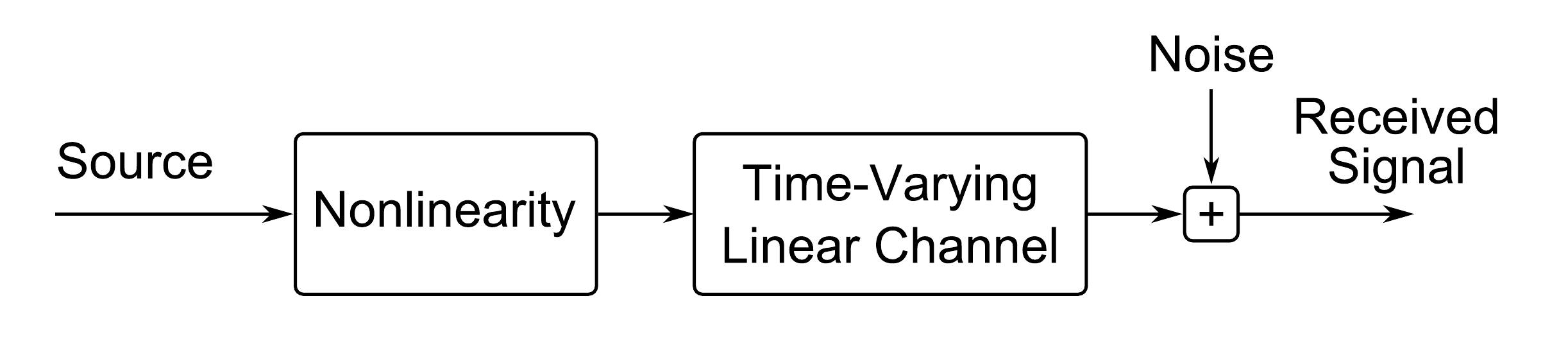}
\caption{The nonlinear channel used in the example consists of a nonlinearity followed by a linear channel.} \LABFIG{diagram_rayleigh}
\end{figure}

Figs.~\ref{fig:results_rayleigh}(a,b) illustrate the tracking results obtained by KRLS-T in these scenarios. As a reference, we include the performance of several state-of-the-art adaptive filtering algorithms, whose Matlab implementations are taken from the Kernel Adaptive Filtering Toolbox, available at \url{http://sourceforge.net/projects/kafbox/}. In particular, we compare KRLS-T with normalized least mean squares (NLMS), extended RLS (EX-RLS), both of which are linear algorithms, see \cite{sayed2003fundamentals}, and quantized kernel LMS (QKLMS) \cite{chen2012quantized}, which is an efficient, kernelized version of the LMS algorithm. A Gaussian kernel $k(\x_i,\x_j)=\exp(-\gamma \|\x_i-\x_j\|^2)$ is used for QKLMS and KRLS-T. In each scenario the optimal hyperparameters of KRLS-T are obtained by performing Type-II ML optimization (see Section \ref{GP_learn} ) on a separate data set of $500$ test samples. The optimal parameters of the other algorithms are obtained by performing cross-validation on the test data set.
%
%
%
To avoid an unbounded growth of the matrices involved in KRLS-T, its memory is limited to $100$ bases which are selected by pruning the least relevant bases (see \cite{vanvaerenbergh2012kernel} for details on the pruning mechanism). The quantization parameter of QKLMS is set to yield similar memory sizes. 
As can be seen in Figs.~\ref{fig:results_rayleigh}(a,b), KRLS-T outperforms the other algorithms with a significant margin in both scenarios. By being kernel-based it is capable to deal with nonlinear identification problems, in contrast to the classical EX-RLS and NLMS algorithms. Furthermore, it shows excellent convergence speed and steady-state performance when compared to QKLMS. Additional experimental comparisons to other kernel adaptive filters can be found in \cite{vanvaerenbergh2012kernel}.

\begin{figure*}[tb]
\centering
\begin{tabular}{cc}
\includegraphics[width=8cm]{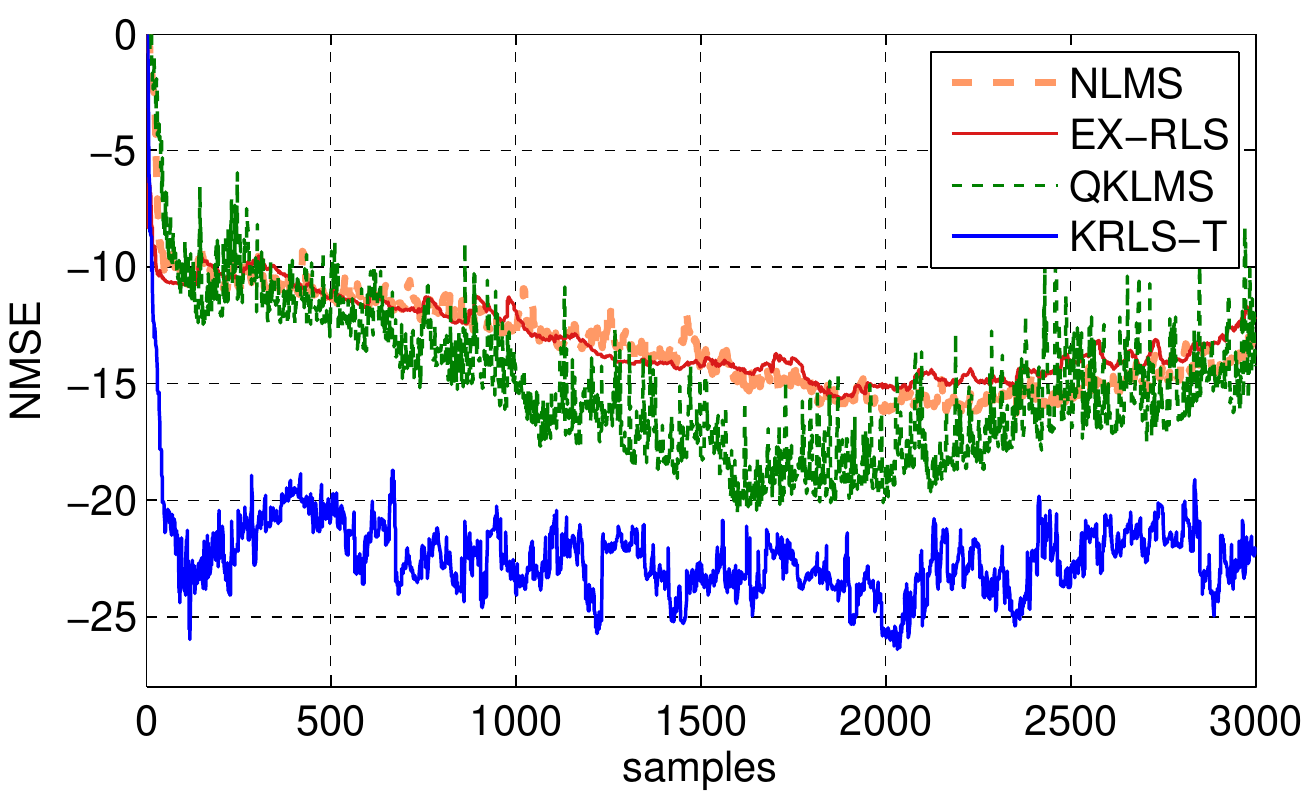} &
\includegraphics[width=8cm]{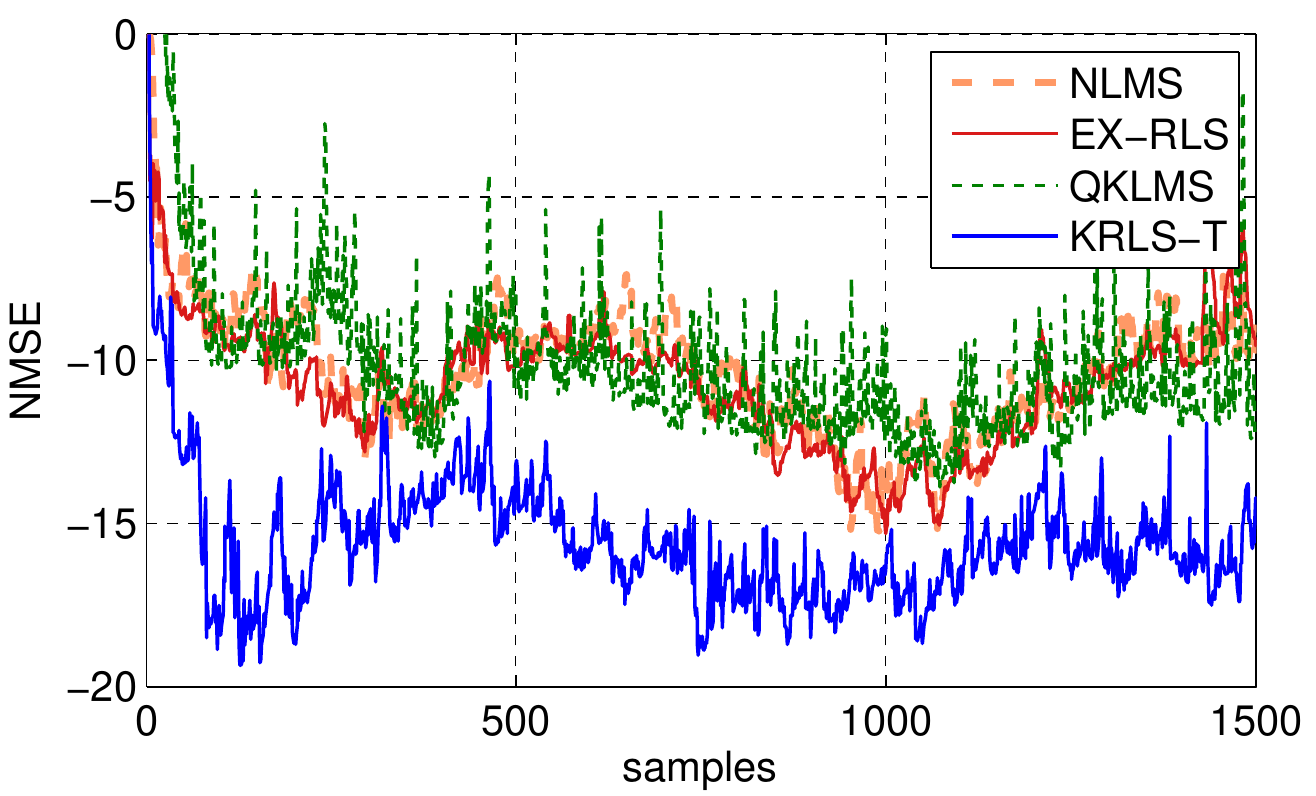}\\
(a) $f_dT = 10^{-4}$, simulated data & (b) $f_dT = 10^{-3}$, simulated data\\
\includegraphics[width=8cm]{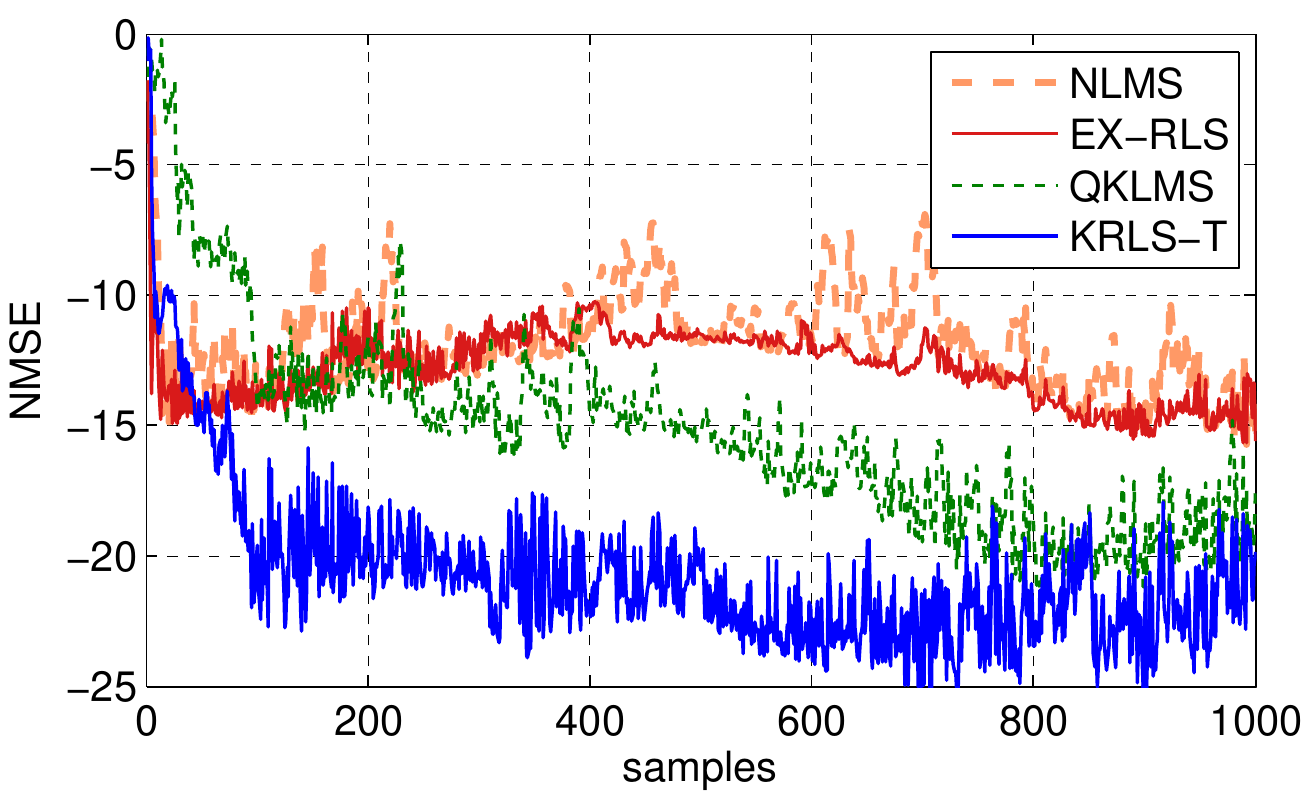} &
\includegraphics[width=8cm]{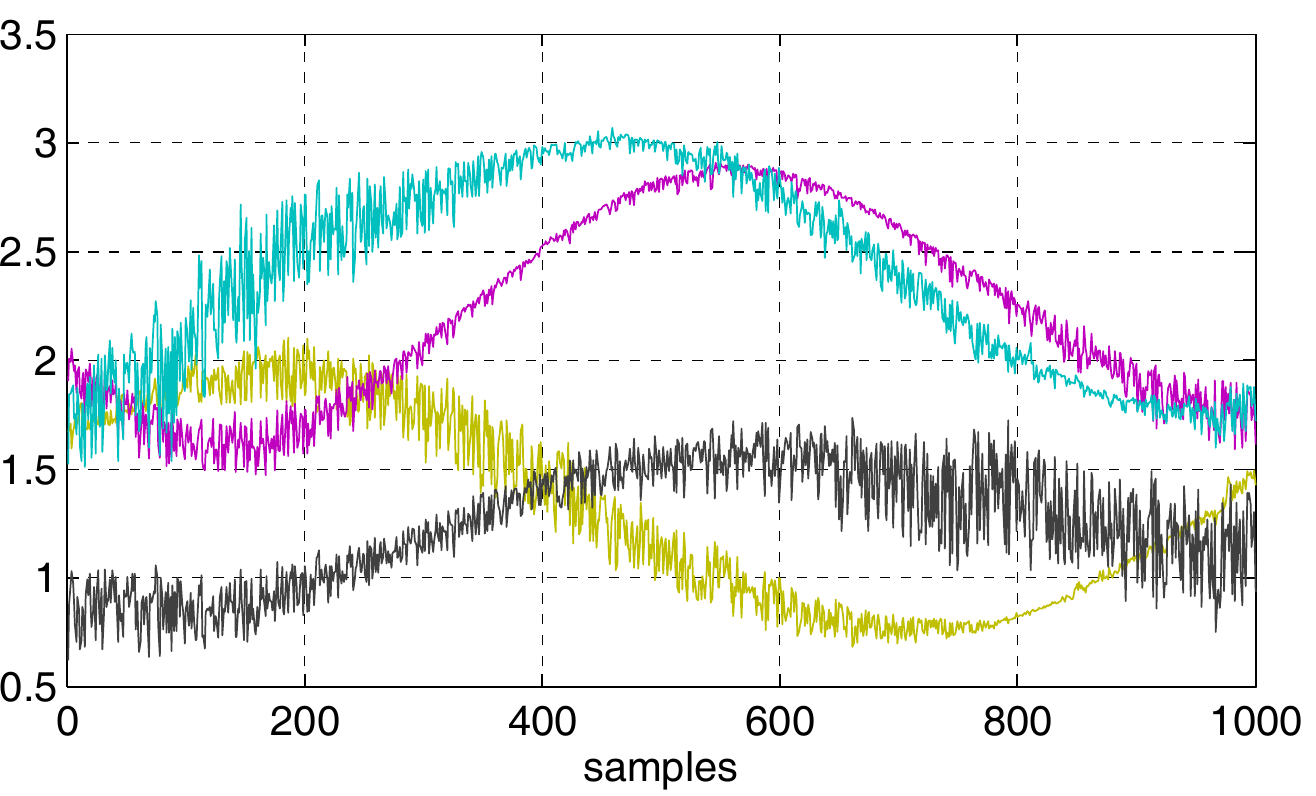} \\
(c) $f_dT = 10^{-3}$, real data & (d) measured linear channels
\end{tabular}
\caption{Tracking results on a nonlinear Rayleigh fading channel: (a) simulation results for a slow-fading scenario; (b) simulation results for a fast time-varying scenario; (c) tracking results on data measured on the test bed with fast time-varying channels; (d) channel taps of the noisy linear channels, measured on the test bed setup.} \LABFIG{results_rayleigh}
\end{figure*}

\begin{table*}
\caption{Steady-state NMSE performance for Fig.~\ref{fig:results_rayleigh}.}%
\newcommand{\zerowidth}[1]{\hbox to 0pt{\hss#1\hss}}
\setlength{\tabcolsep}{.9em}
\begin{tabular}[C]{| r | c | c | c | c |} \hlx{hv]}
\multicolumn{1}{|c|}{\zerowidth{}} & %
\multicolumn{1}{c|}{NLMS} & %
\multicolumn{1}{c|}{EX-RLS} & %
\multicolumn{1}{c|}{QKLMS} & %
\multicolumn{1}{c|}{KRLS-T} \\ \hlx{vhv}
$f_dT=10^{-4}$, simulated & $-13.3$ dB & $-13.0$ dB & $-14.6$ dB & $-22.3$ dB\\
$f_dT=10^{-3}$, simulated & $-10.6$ dB & $-11.0$ dB & $-9.9$ dB & $-15.3$ dB\\
$f_dT=10^{-3}$, real data & $-11.5$ dB & $-12.5$ dB & $-15.8$ dB & $-21.3$ dB\\
\hlx{vhs}
\end{tabular}
\label{table:tracking_nmse}
\end{table*}

In a second setup we used a wireless communication test bed that allows to evaluate the performance of digital communication systems in realistic indoor environments. This platform is composed of several transmit and receive nodes, each one including a radio-frequency front-end and baseband hardware for signal generation and acquisition. The front-end also incorporates a programmable variable attenuator to control the transmit power value and therefore the signal saturation. A more detailed description of the test bed can be found in \cite{gutierrez2011frequency}. Using the hardware platform, we reproduced the model corresponding to Fig. \ref{fig:diagram_rayleigh} by transmitting clipped orthogonal frequency-division multiplexing (OFDM) signals centered at 5.4 GHz over real frequency-selective and time-varying channels. Notice that, unlike the simulated setup, several parameters such as the noise level and the variation of the channel coefficients are unknown. To have an idea about the channel characteristics, we first measured the indoor channel using the procedure described in \cite{gutierrez2011frequency}. As an example, the variation of the four main channel coefficients is depicted in Fig.~\ref{fig:results_rayleigh}(d), indicating a normalized Doppler frequency around $f_dT=10^{-3}$. We then transmitted periodically OFDM signals with the transmit amplifier operating close to saturation and acquired the received signals. The transmitted and received signals were used to track the nonlinear channel variations as in the simulated setup. The results, shown in Fig.~\ref{fig:results_rayleigh}(c), are similar to those of the simulated setup. Finally, the steady-state NMSE performances of all three scenarios, Figs.~\ref{fig:results_rayleigh}(a,b,c), are summarized in Table~\ref{table:tracking_nmse}.


\section{Gaussian Processes for Classification}\label{gpc}

For classification problems, the labels are drawn from a finite set and GPs return a probabilistic prediction for each label in the finite set,  i.e., how certain is the classifier about its prediction. In this tutorial, we limit our presentation of GPs for classification (GPC)
for binary classification problems, i.e., $\y_{i}\in\{0,1\}$. For GPC, we change the likelihood model for the latent function at $\x$ using a \emph{response function}
$\Phi(\cdot)$:
\begin{equation}\LABEQ{like_ind_class}
p(\y=1|\fv(\x)) = \Phi(\fv(\x)).
\end{equation}
The response function ``squashes'' the real-valued latent function
to an $(0,1)$-interval that represents the posterior probability for
$\y$ \cite{rasmusssen2006gaussian}. Standard choices for the response function
are $\Phi(a)=1/(1+\exp(-a))$ and
the cumulative density function of a standard normal distribution, used in logistic and probit regression respectively.

The integrals in \EQ{6r} and \EQ{5r} are now analytically
intractable, because the likelihood and the prior are not
conjugated. Therefore, we have to resort to numerical methods or
approximations to solve them. The posterior distribution in
\EQ{posteriorf} is typically single-mode and the standard methods
approximate it with a Gaussian \cite{rasmusssen2006gaussian}.  Using a Gaussian
approximation for \EQ{posteriorf} allows exact marginalization in
\EQ{5r} and we can use numerical integration for solving \EQ{6r}, as
it involves marginalizing a single real-valued quantity. The two
standard approximations are the Laplace method or expectation propagation (EP)
\cite{Minka01}. In \cite{Kuss05}, EP is shown to be a more
accurate approximation.


\subsection{Probabilistic channel equalization}

GPC predictive performance is similar to other nonlinear discriminative methods, such as SVMs. However, if the probabilistic output is of importance, then GPC outperforms other kernel algorithms, because it naturally incorporates the confidence interval in its predictions. In digital communication, channel decoders follow equalizers, which work optimally when accurate posterior estimates are given for each symbol. To illustrate that GPC provide accurate posterior probability estimates, we equalize a dispersive channel model like the one in \FIG{diagram_rayleigh} using GPC and SVM with a probabilistic output. These outputs are subsequently fed to a low-density parity-check (LDPC) belief-propagation based channel decoder to assess the quality of the estimated posterior probabilities. Details for the experimental set up can be found in \cite{olmos2010joint} in which linear and nonlinear channel models are tested. We now summarize the results for the linear channel model in that paper.

In \FIG{gpc}, we depict the posterior probability estimates versus the true posterior probability, in (a) for the GPC-based equalizer and in (b) for SVM-based equalizer, to emphasize the differences between the equalizers we use a highly noisy scenario with normalized signal-to-noise ratio of $2$ dB. If we threshold at 0.5, both equalizers provide similar error rates and we cannot tell if there is an advantage from using GPC. However, if we consider the whole probability space, GPC predictions are significantly closer to the main diagonal that represents a perfect match, hence GPC provides more accurate predictions to the channel decoder.

To further quantify the gain from using a GPC-based equalizer with accurate posterior probability estimates, we plot the bit error rate (BER) in \FIG{GPCBER} after the probabilistic channel encoder, in which the GPC-based equalizer clearly outperforms the SVM-based equalizer and is close to the optimal solution (known channel and forward-backward (BCJR) equalizer). This example is illustrative of the results that can be expected from GPC when a probabilistic output is needed to perform optimally.


\begin{figure*}[tb]
\centering
\begin{tabular}{cc}
\includegraphics[width=8cm]{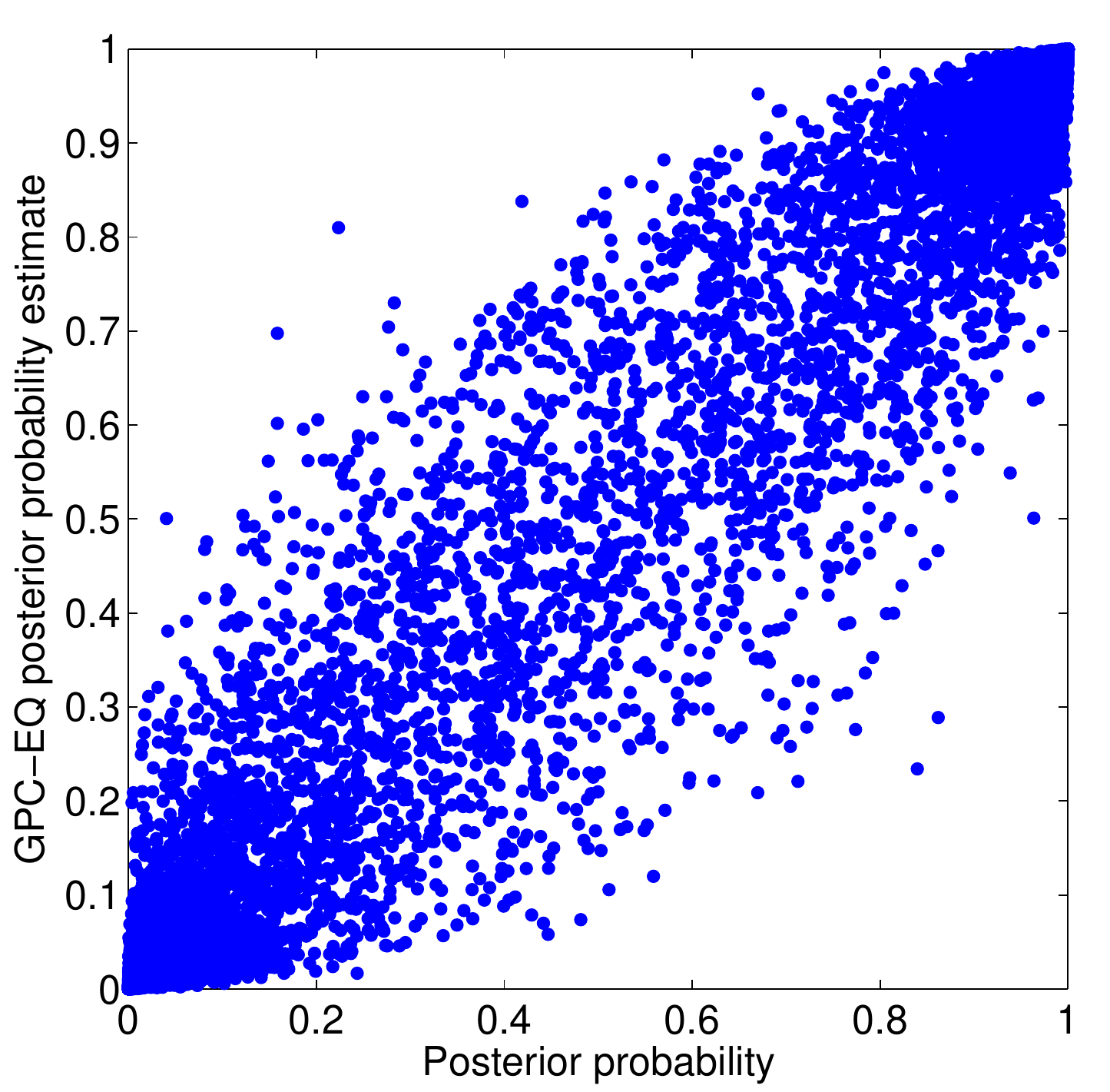} &
\includegraphics[width=8cm]{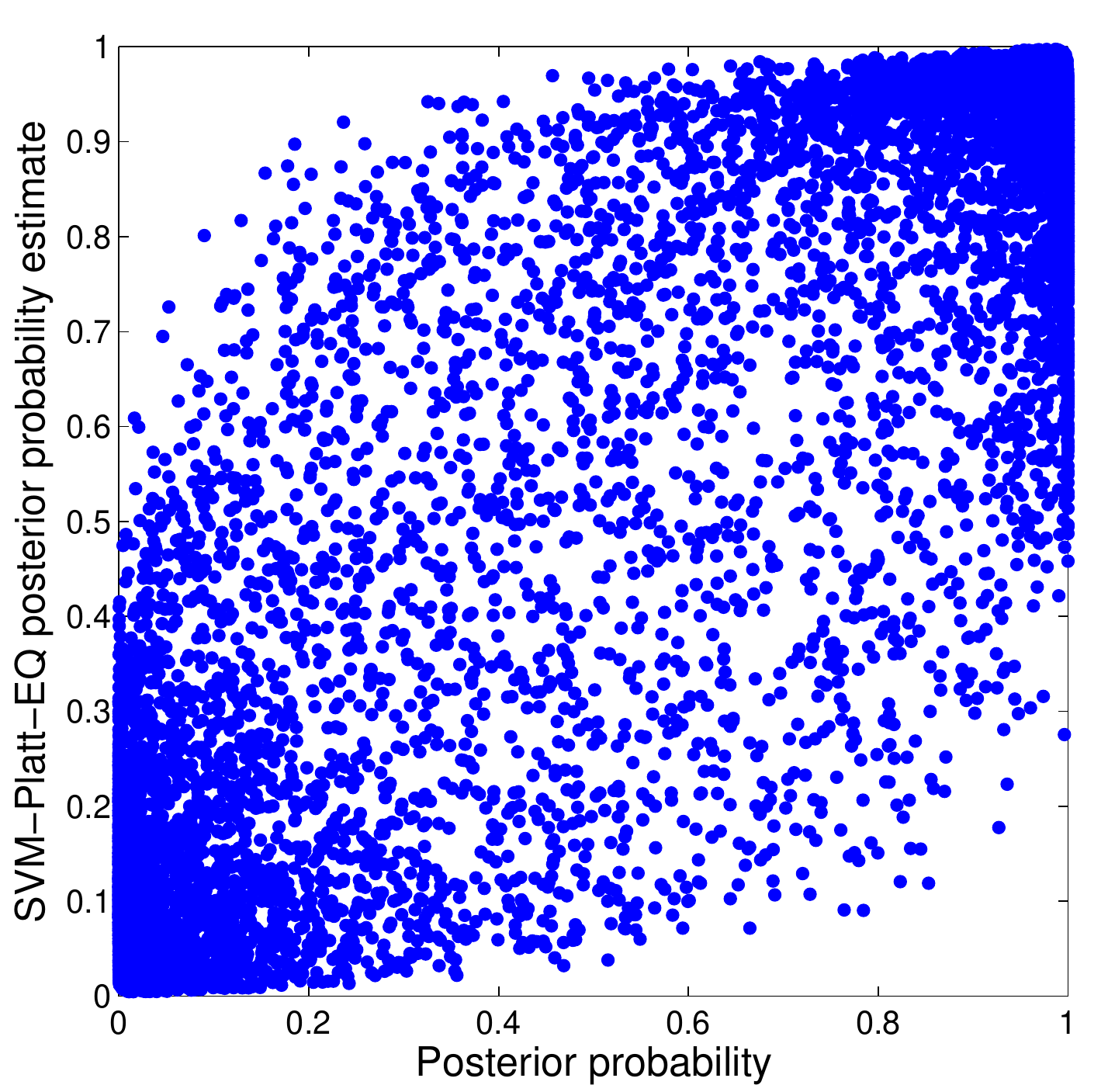} \\
(a)  & (b)
\end{tabular}
\caption{GPC as probabilistic channel equalizer: (a) calibration curve for the GPC and (b) calibration curve for the SVM.} \LABFIG{gpc}
\end{figure*}

\begin{figure}[tb]
\centering
\includegraphics[width=8cm]{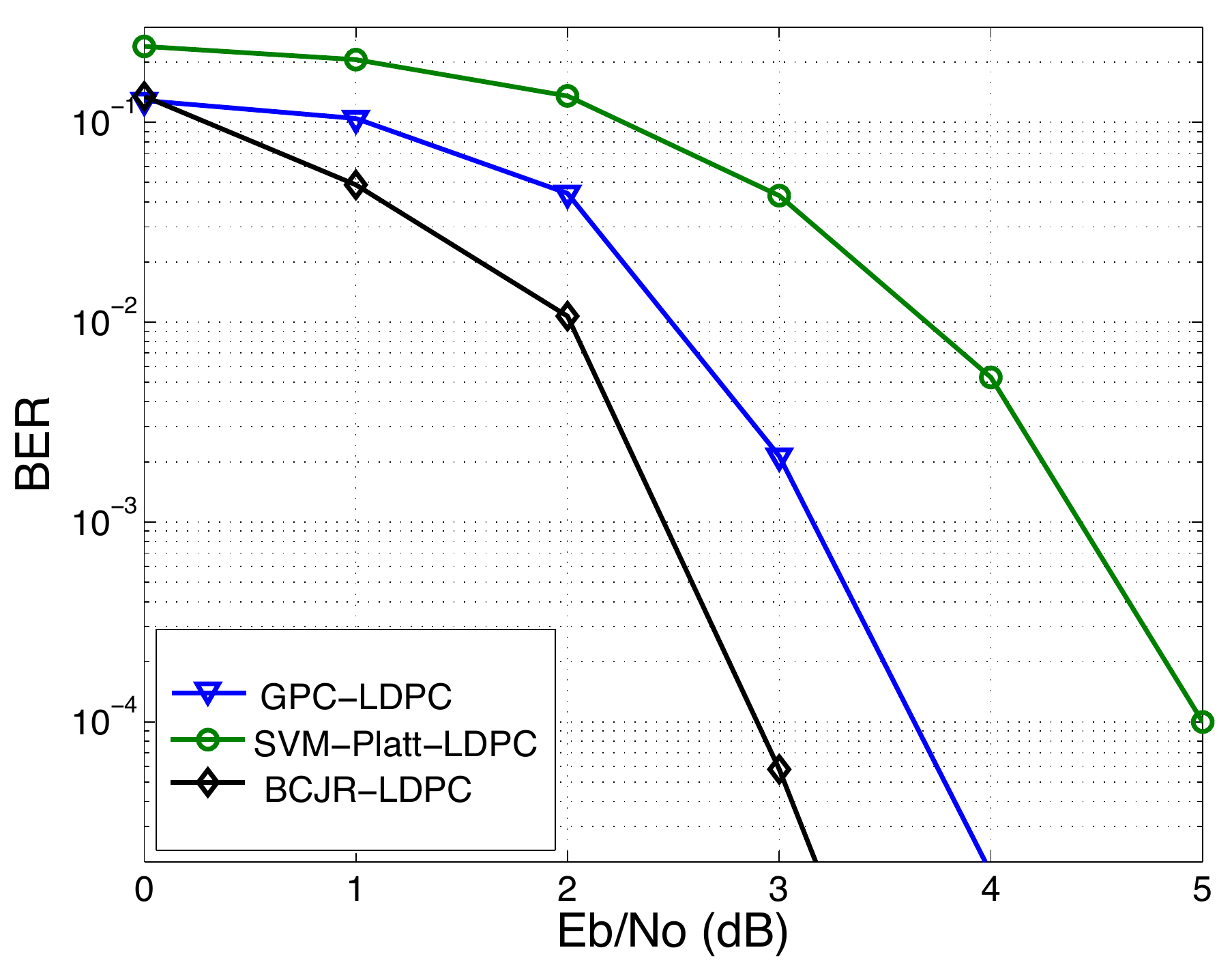}\\
\caption{GPC and SVM as probabilistic channel equalizer in channel LDPC decoding: BER for the GPC-LDPC ($\triangledown$), the SVM-LDPC ($\circ$) and the optimal solution ($\diamond$).  } \LABFIG{GPCBER}
\end{figure}



\section{Discussion}

In this tutorial, we have presented  Gaussian Processes for Regression in detail from the point of view of MMSE/Wiener filtering, so it is amenable to signal processing practitioners. GPR provides the same mean estimate as KLS or KRR for the same kernel matrix. On the plus side, GPR provides error bars that take into account the approximation error and the error from the likelihood model, so we know the uncertainty of our model for any input point (see \FIG{gpr} ), while KLS assumes the error bars are given by the likelihood function (i.e., constant for the whole input space). Additionally, GPR naturally allows computing the hyper-parameters of the kernel function by sampling or maximizing the marginal likelihood, being able to set tens of hyper-parameters, while KLS or SVM need to rely on cross-validation, in which only one or two parameters can be easily tuned. On the minus side, the GP prior imposes a strong assumption on the error bars that might not be accurate, if the latent variable model does not follow a Gaussian process. Although, in any case, it is better than not having error bars.

We have also shown that some of the limitations of the standard GPR can be eased. GPs can be extended to non-Gaussian noise models and classification problems, in which GPC provides an accurate a posteriori probability estimate. The computational complexity of GPs can be reduced considerably, from cubic to linear in the number of training examples, without significantly affecting the mean and error bars prediction. Finally, we have shown the GP can be solved iteratively, with an RLS formulation that can be adapted to non-stationary environments efficiently.

Instead of covering more methods and applications in detail, our intention was to provide a tutorial paper on how to use GPs in signal processing, with a number of illustrative examples. Nevertheless, since we assume that there are several other methods and applications that are relevant to the reader, we finish with a brief list of further topics. In particular, GPs have also been applied to problems including modeling human motion \cite{Wang08}, source separation \cite{Liutkus11}, estimating chlorophyll concentration \cite{Pasolli10}, approximating stochastic differential equations \cite{Archambeau07} and multi-user detection \cite{Murillo09}, among others.

\section{Acknowledgments}
The authors would like to thank Jes\'us Guti\'errez, University of Cantabria, Spain, for his assistance in capturing the data of the test bed experiment.

\bibliographystyle{IEEEtran}
\bibliography{biblio}

\end{document}